\def\year{2023}\relax
\documentclass[letterpaper]{article} 
\usepackage{aaai22}  
\usepackage{times}  
\usepackage{helvet}  
\usepackage{courier}  
\usepackage[hyphens]{url}  
\usepackage{graphicx} 
\usepackage{natbib}  
\usepackage{caption} 
\DeclareCaptionStyle{ruled}{labelfont=normalfont,labelsep=colon,strut=off} 
\usepackage{algorithm}
\usepackage{algorithmic}

\usepackage{newfloat}
\usepackage{listings}
\floatstyle{ruled}
\newfloat{listing}{tb}{lst}{}
\floatname{listing}{Listing}

\usepackage{todonotes}
\usepackage{multirow}
\usepackage{caption}
\usepackage{subcaption}
\usepackage{colortbl}
\usepackage{graphicx}
\usepackage{algorithm}
\usepackage[hidelinks]{hyperref}
\usepackage[multiple]{footmisc}

\title{The Five-Dollar Model:\\ Generating Game Maps and Sprites from Sentence Embeddings}
\author{
    Timothy Merino, Roman Negri, Dipika Rajesh, M Charity, Julian Togelius
}
\affiliations{
    Game Innovation Lab\\ New York University - Tandon School of Engineering\\
    Brooklyn, New York\\
    tm3477@nyu.edu, rvn9303@nyu.edu, dipika.rajesh@gmail.com, mlc761@nyu.edu, julian@togelius.com
}

\begin{document}

\maketitle

\begin{figure*}
    \centering
    \includegraphics[width=\linewidth]{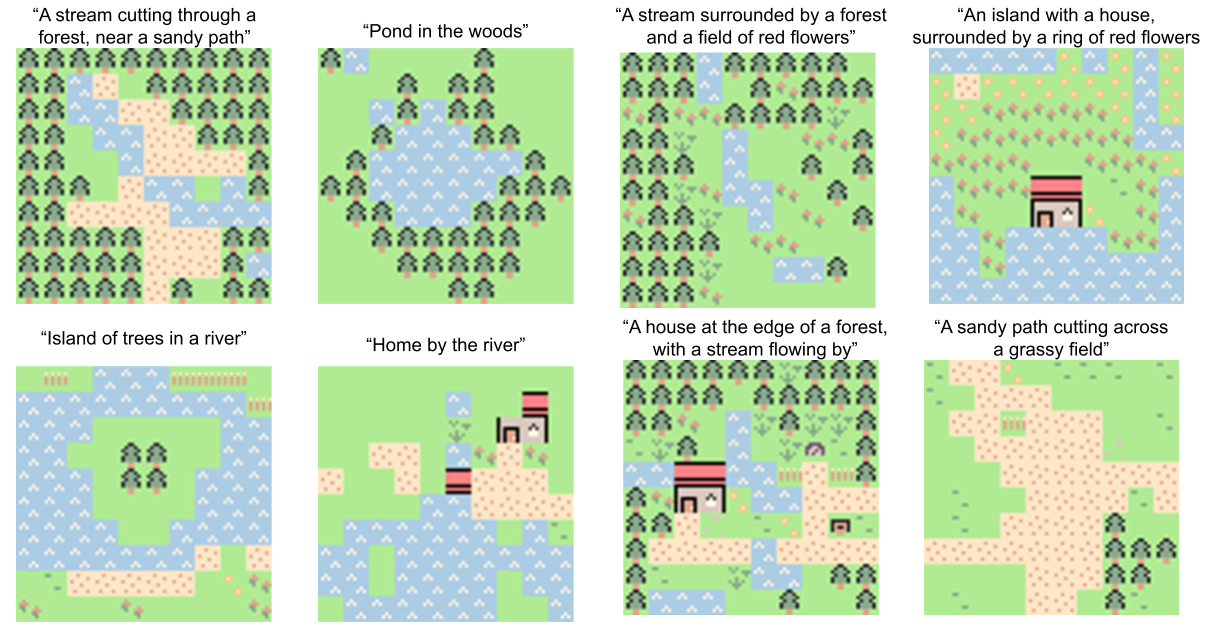}
    \caption{New video game maps generated using unseen text prompts. }
    \label{fig:MainFigMaps}
\end{figure*}

\begin{figure}[ht!]
    \centering
    \includegraphics[width=\linewidth]{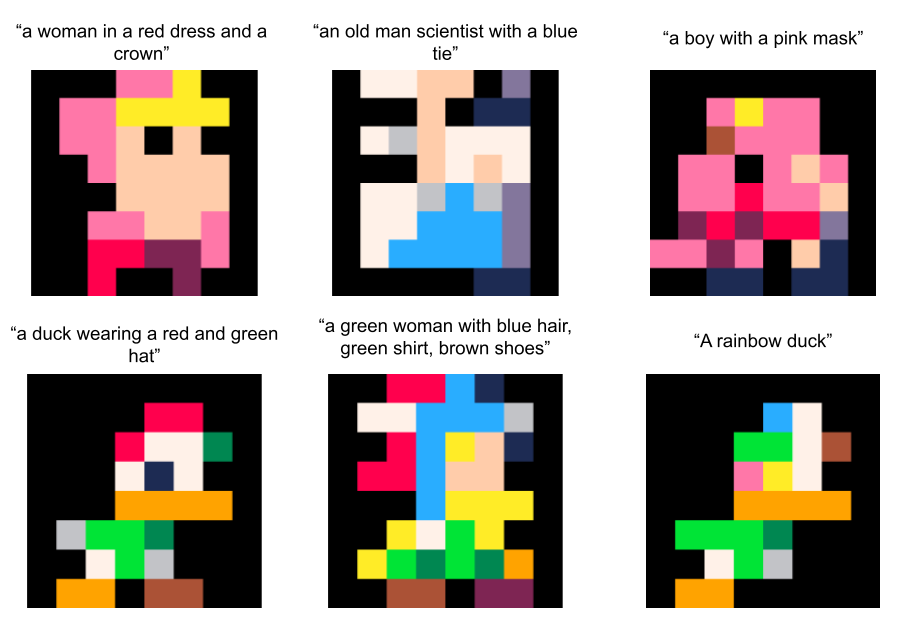}
    \caption{New sprite characters generated using unseen text prompts. }
    \label{fig:MainFigSprite}
\end{figure}

\begin{figure}[ht!]
    \centering
    \includegraphics[width=\linewidth]{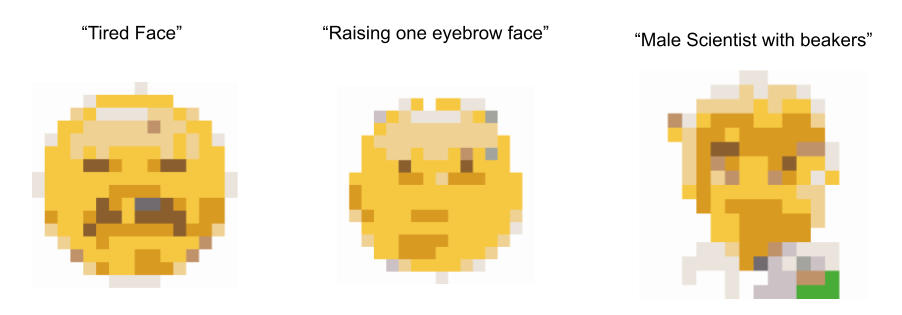}
    \caption{New emojis generated using unseen text prompts. }
    \label{fig:MainFigEmoji}
\end{figure}

\begin{abstract}

The five-dollar model is a lightweight text-to-image generative architecture that generates low dimensional images from an encoded text prompt. This model can successfully generate accurate and aesthetically pleasing content in low dimensional domains, with limited amounts of training data. Despite the small size of both the model and datasets, the generated images are still able to maintain the encoded semantic meaning of the textual prompt.  We apply this model to three small datasets: pixel art video game maps, video game sprite images, and down-scaled emoji images and apply novel augmentation strategies to improve the performance of our model on these limited datasets. We evaluate our models performance using cosine similarity score between text-image pairs generated by the CLIP VIT-B/32 model.

\end{abstract}

\section{Introduction}

Recent developments with text-to-image generators in the AI community have spurred a renaissance of large generative models. Both open and closed source generators, such as DALL-E and StableDiffusion have demonstrated their versatility and range in image generation, from hyper-realistic photos to abstract artworks. However, to generate these quality images, these large models require large amounts of training data---both in text captions and paired images---large amounts of compute and training time, and large amounts of storage. Though these models boast incredible textual understanding and image fidelity, they cannot be used in every domain. For more specific tasks, such as creating assets like sprites, textures, or levels for video games, these models require extensive prompt engineering or fine tuning to meet domain-specific constraints.

Within procedural content generation (PCG) research, various smaller models have been proposed specifically to create assets. These models are much simpler in architectural design and include evolutionary models, GANs, VAEs, and others. However, most PCG models lack conditional prompting. Conditional prompting is a technique used to guide the generation process towards more specific artifacts using a provided condition, such as a class label or text description. A small-scale model that requires low amounts of training data and has direct text-to-image generation capabilities would benefit game developers considerably, as it would facilitate content creation and provide a greater degree of control over the output of the AI system.

We introduce the five-dollar model\footnote{A model so simple, one of the authors bet another \$5 it wouldn't work.}---a text-to-image generator designed for low-dimensional, pixel-size images and categorical-based samples. We use this model to generate game levels (Figure \ref{fig:MainFigMaps}), sprites (Figure \ref{fig:MainFigSprite}), and emoji images (Figure \ref{fig:MainFigEmoji}). We find that despite the conceptual simplicity and comparatively small size of our model, it demonstrates surprising performance and language understanding in these domains. This generator is remarkably fast because of its small size and lack of iterative steps, making it suitable for real-time content generation. It is also easy to train because of the lack of adversarial dynamics (unlike GANs) and it can be trained using relatively small training sets. Because of the lack of denoising (unlike diffusion models), it is suitable for discrete and categorical content such as tile-based maps. Finally, the simplicity of the model makes it easier to study and potentially modify after training.

\section{Background}

Procedural content generation in games is an active field of game AI research. This field includes algorithmically creating artificial samples for game content, such as levels, sprites, or narratives \cite{pcgbook}. Previous works have used evolutionary algorithms \citep{charity2022aesthetic}, generative adversarial networks (GANs) \citep{volz2018evolving}, variational autoencoders (VAEs) \citep{sarkar2021generating}, VQ-VAEs \citep{saravanan2022pixel}, cellular automata \citep{earle2022illuminating}, diffusion models \citep{xu2023dream3d}, transformer models \citep{todd2023level}, and more.

Conditional generation allows for more control over the output of the black-box neural network models by providing auxiliary information, called the condition. Conditional information can come in many forms. Class labels \cite{DBLP:journals/corr/MirzaO14}, text descriptions \cite{pmlr-v48-reed16}, and other images \cite{dumoulin2017a} have been used as conditional inputs to guide image-generation models towards a specific goal. Works done by Van den Oord \citep{van2016conditional}, Kang \citep{kang2020contragan}, Mirza \citep{DBLP:journals/corr/MirzaO14}, Perez \citep{Perez2017FiLMVR}, and Isola \citep{isola2017image} with conditional generators on various domains such as face and image generation demonstrate the effectiveness of control over the final output when using conditional inputs all the while retaining quality. 

Text-to-Image generation employs natural language image descriptions as a means to condition generative image models. This class of models aims to generate high quality images that semantically align with the provided input prompt. With the rise of conditional diffusion \citep{sinha2021d2c} and Stable Diffusion models \citep{ramesh2022hierarchical}, text-to-image generation has seen a huge increase in popularity. Current research in this domain focuses largely on the zero-shot capabilities of such models --- the ability for them to generalize to unseen data. 

Procedural Content systems are most often designed for a specific game, where zero-shot generalization is not a primary concern.  With this paper, we aim to explore domain-specific text to image generation in the context of digital game art. In contrast to state of the art systems, which are typically trained on large scale datasets of real world RGB images, we train on low dimensional categorical data.

\section{Methodology}

\subsection{Model Architecture}
The five-dollar text-to-image generator model is a simple feedforward network. The model learns a mapping from the embedded text vector to a higher dimension categorical image. The generator model has 2 inputs: (1) an embedding vector representing the text condition and (2) a random noise vector. The input text caption is encoded outside of the system by a pre-trained sentence transformer model. We use the multi-qa-MiniLM-L6-cos-v1 sentence transformer model, which encodes text into a 384-dimensional vector. The noise layer allows for diversity in the final generated output with the same text encoding. The two input layers are combined via a concatenation layer, then reshaped by a dense layer and fed into the network. This reshaped input is then passed through residual blocks --- 2 Convolutional layers with Batch Normalization --- to generate an image. Upsampling layers are placed before the first residual blocks, to reach the desired output shape (8x8 for sprites, 10x10 for maps, 16x16 for emojis). 

We experiment with different numbers of residual blocks, kernel sizes, and convolutional filters within each dataset. We choose the architecture that maximizes validation accuracy, with some consideration for model size, to generate the final images.

\begin{figure}[ht!]
    \centering
    \includegraphics[width=\linewidth]{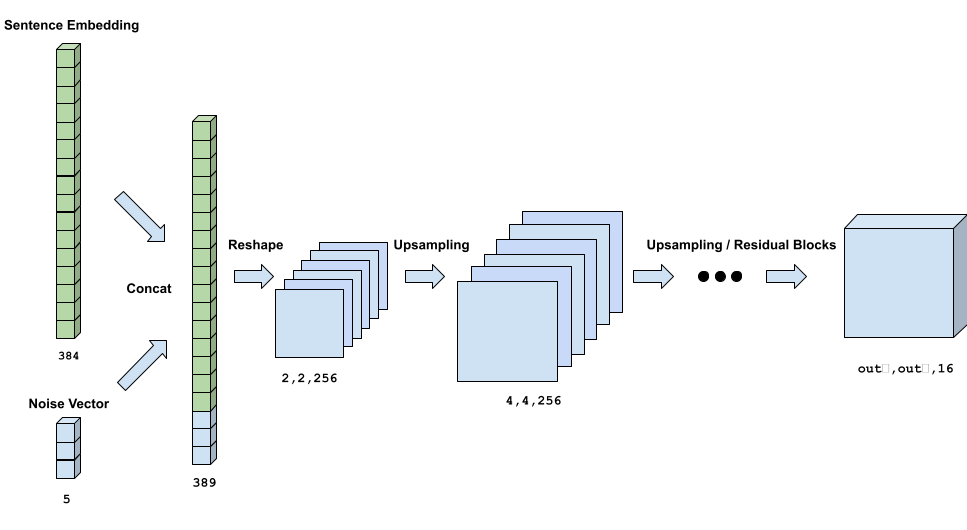}
    \caption{Architecture design of the five-dollar model. This outputs a NxNx16 image. For maps, N = 10. For sprites, N=8. For emojis, N=16 }
    \label{fig:model_architecture}
\end{figure}

Figure \ref{fig:model_architecture} shows the model architecture of the five-dollar model. Further experimentation was done with more complex architectures involving Conditional Instance Normalization \cite{dumoulin2017a} and Feature-wise Linear Modulation \citep{Perez2017FiLMVR} conditioning layers. The results of these experiments are shown in Table \ref{tab:appendix_table} of the appendix. In general, these conditioning methods did not noticeably improve the generated output compared to the input concatenation-based conditioning. We instead focus on the simpler concatenation-conditioned model throughout the paper.

\subsection{Training Data}
We train our models in a supervised setting, using pairs of caption embeddings and images. We use categorical cross-entropy loss between the generated image and training image, using an  \textsc{Adam} optimizer with learning rate = $0.001$

\begin{figure}[ht!]
    \centering
    \includegraphics[width=\linewidth]{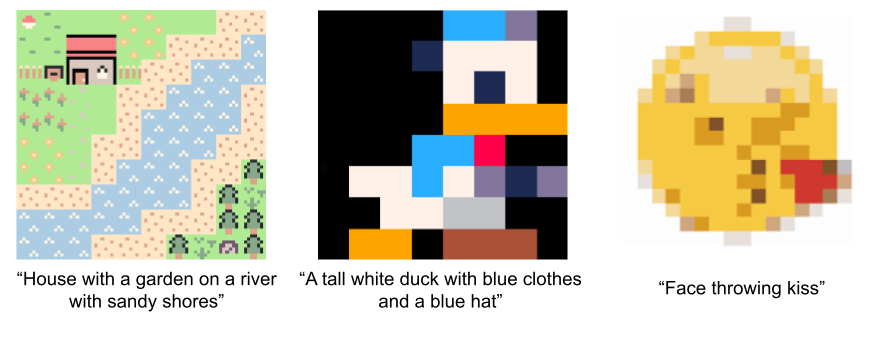}
    \caption{Example images with labels from all 3 datasets. From left to right: Map, Sprite, Emoji. }
    \label{fig:dataset_examples}
\end{figure}

We use three datasets to evaluate our model --- the largest dataset containing 882 samples and the smallest containing 100 samples. These datasets are selected due to their small pixel resolutions and diverse domains, ultimately allowing us to demonstrate the range of output for the model with the text conditioned generation. In contrast to most image datasets, ours use a one-hot encoded representation rather than RGB. This is similar to functional constraints present in many games, where assets are limited by color space or tiles. 

The first is a crowd-source map dataset, consisting of 882 images that resemble sections of the game map from retro RPGs such as Pokemon. These maps are represented by a one-hot encoded 10x10x16 image, using 16 possible 8-pixel tiles. Each map is paired with a sentence describing the scene depicted in the map, such as "A house in a grassy field".  452 of the maps in this dataset come from existing work, without labels \citep{charity2022aesthetic}. Labels were manually added to these existing maps by the authors. The remaining 430 caption-map pairs were created via an interactive webpage, which was shared via the authors' Twitter.  Users were asked to create a map using the tile set, and provide one sentence describing the map they created.

The second dataset consists of emoji images and labels \cite{hendrycks2018robustness}. These 10,000 emoji images are represented by 32x32x3 RGB images. They are separated by style (Apple, Facebook, etc), and are clustered into 50 categories, such as Face, Animal, and Bird. We train our models on a subset of this dataset, consisting of Apple-style emojis in 29 categories. This filtering results in 663 emoji images. We downscale these images by a factor of 2 using inter-area interpolation, to obtain 16x16x3 RGB images. Then, 16 colors are selected using K-Means clustering of color values across the filtered Apple-style dataset. Each pixel is then quantized based on this color palette, resulting in a 16x16x16 one-hot encoded image.

The third dataset uses 100 8x8 pixel sprites created from the PICO-8 game engine - using 16 possible colors according to the PICO-8 palette. These sprites were designed based on recognizable game and cartoon characters (i.e. Super Mario, Adventure Time, and the Simpsons) but with their own unique designs and structures. Instead of labeling these characters by their canonical names, we manually labeled them with generic descriptions. For example, instead of the label 'Mario' for the sprite, it is described as 'a man with a mustache dressed in red.' 

Figure \ref{fig:dataset_examples} shows example caption-image pairs from each dataset. The 16 tile and color values for each dataset can be found in Figure \ref{fig:palettes} in the appendix.

\subsection{Encoding Augmentations}

To enrich the text embedding space, we use 4 different methods of augmentation --- (1) using GPT-4 to generate alternate labels for each image, (2) increasing the dataset with multiplicative Gaussian noise, and (3)/(4) two different adaptations of MixUp \citep{zhang2017mixup}, to linearly interpolate between samples. Each encoding augmentation is discussed in further details below.

In order to create GPT-4 generated alternate labels, we pass our existing dataset labels along with a custom prompt into GPT-4, tasking it to write an alternate version of each label using simple language. We instruct GPT to assume it has the vocabulary of a 10-year old, as early experiments resulted in uncommon words that were out of distribution with our original dataset. The prompt used for the map dataset can be found in the Appendix in Figure \ref{lst:gpt_prompt}.
  
The alternate labels are then passed through the text encoder model to get new embeddings vectors. We only generate 1 alternate version of each label, as it was challenging to generate more alternate versions without duplicating labels.

We also incorporate a multiplicative Gaussian noise augmentation. For an arbitrary sentence embedding vector in our dataset, we generate a 384-dimensional noise vector, with each element $\in \mathcal{N} ~ (1, 0.15)$. By taking the Hadamard product of the sentence embedding vector and the noise vector, a "noisy" embedding is generated. For our final models, we use a dataset with 3 noisy embeddings per original label.

We also experiment with two forms of MixUp. In the first form,  which we call "alt label interpolation", we generate $n$ vectors interpolated between an embedded label and its matching embedded GPT label. In the second form, random Mixup, we generate a vector through interpolation between a sample and another randomly selected sample, with $\lambda = 0.5$\footnote{This is the degree to which each sample will be represented in the interpolated vector. A value of $.5$ indicates equal representation of each input vector.}. 

Both the GPT alt labels and noise augmentations improved the performance of our models in all 3 datasets. We only saw improvements to the models from the two Mixup forms in the map dataset. In the other datasets, while alt label interpolation did not cause noticable performance degredations, random MixUp resulted in much poorer generated sprites.

\section{Experiments}

We create various experiments to evaluate the language understanding and generative abilities of the five-dollar model in each domain.

First, the latent space of the model is investigated  using interpolation, also known as ``walking the latent space" \cite{radford2015unsupervised}. A smooth latent space should result in a gradual transition of the intermediate outputs between the two associated images. We manually select two points in the dataset with either similar semantic meanings, or similar image representations for these experiments.

Secondly, to evaluate the natural language understanding of our model, we use latent vector arithmetic. Using arithmetic operations on the encoded text vectors of two similar images, we attempt to isolate vectors which represent important semantic concepts within the image domain. For example, by subtracting ``neutral face" from ``angry face" in the emoji domain, we obtain a vector that represents ``anger". We can then visualize adding this vector to other text prompts, to see whether the model encodes an intuitive understanding of these concepts.

\section{Model Training}
We lay out the entire training pipeline in 4 steps: (1) Alternate Label Creation with GPT-4 (2) Data Preparation, (3) Data Augmentation, (4) Model training. Step (1), the alternate label step consists of generating the alternate labels from GPT-4 in order to increase the base size of the dataset. Step (2), the data preparation step, includes quantization of images into one-hot encoded representation and encoding of text labels using the embedding model. Step (3), applying the augmentations mentioned above such as MixUp and noise augmentations. Finally, step (4), the model training stage, which in our case included a grid search for the best performance in each domain.

At inference time the only models used are the pretrained embedding model and the five-dollar model, completely foregoing of steps 1 and 3. Large models, such as GPT-4, are only used during the alternate label creation step. Thus, GPT-4 gives no increase in computational resources required to use this model in downstream applications.

\section{Evaluation}

To fully understand the capabilities of these models we aimed to optimize the following hyperparameters through a grid search: the noise input vector's dimension, the filter counts, sizes of the kernels, the number of residual blocks, and the conditioning type\footnote{The three options for conditioning types are concatenation based (depicted in Figure \ref{fig:model_architecture}), Conditional Instance Normalization (CIN layers after every residual block), or Feature-wise Linear Modulation (FILM layers after every residual block).}.
The models were evaluated based on validation accuracy within their domains with validation sets consisting of withheld human-made labels and a matching pixel map, sprite, or emoji respectively for each of the domains. 

While a correlation between validation accuracy and output quality is noted, the augmentation strategies used on the dataset introduced some information leakage between the training and validation sets. To combat this, we evaluate the top performing models in two ways. \footnote{When deciding between two models with similar performance, the model with the lower number of parameters was selected. } 

First, we subjectively evaluate the top models' performance on various sets of unseen labels. These were designed to test the semantic understanding of the models, such as applying a different color to a learned outline i.e.:\textit{ duck with a pink shirt } as the unseen sprite, with duck, pink, and shirt all having independent examples in the training set. More examples of unseen prompts can be found in Figure \ref{lst:unseen_listings} in the Appendix. 

Secondly, we use CLIP Score as a measure of the text-image alignment. Contrastive Language-Image Pre-Training (CLIP) models embed similar text labels and images into an overlapping embedding space, aiming to maximize the cosine similarity of the embedded vectors for corresponding text-image pairs \cite{radford2021learning}. We use a pretrained \textit{clip-vit-base-patch32} as the CLIP model in this paper. A scaled version of this similarity metric, called the CLIP Score, has a high correlation with human judgements \cite{hessel-etal-2021-clipscore}. We compute the Clip Score as the scaled cosine similarity, with a scaling factor of $2.5$. 

We utilize this metric as an indicator of model performance without the need for human review. In contrast to some state-of-the-art generative models, our model does not include direct optimization for text-image alignment when training, maintaining the lightweight characteristics of the model. However, CLIP Score has limitations as a metric for our model evaluation. Due to the low dimensionality of our data, the domains of our images diverge significantly from the datasets used to train CLIP. Therefore, knowing CLIP performs well on zero shot tasks, and CLIP Score demonstrates its ability to reason about non-photographic art, we perform experiments to validate its use in our domains.

For each domain, we generate a fake dataset by pairing real captions with randomly tiled images\footnote{These were generated through a simple RandInt() function in the needed shape to match the real data.}. We then compare the average CLIP Score of the real dataset vs the fake, random-tile dataset. We determine the validity of this metric by comparing these two scores: A significantly higher score on the real dataset vs the random dataset points to some semantic understanding of the pretrained CLIP model in our domains.

Lacking performance was noted when prompting CLIP with only the raw text input given to the generator models. Therefore, to gain a more representative metric we prepend the text with a dataset specific preprompt to guide the CLIP model towards our domain-specific task. The preprompts for each domain can be found in Figure \ref{lst:preprompt_listings} in the Appendix.

Now that we have target CLIP Score for our generators and after finding candidate hyperparameters via grid search, we compute CLIP Score of each model using a set of unseen labels. Table \ref{tab:model_clipscore} shows the Clip Scores of the top standard and non-standard models in each domain\footnote{''Standard" refers to models without added CIN or FiLM conditioning}. 

\section{Results}
We find that the five-dollar Model was able to generate aesthetically pleasing images that were semantically consistent with the input prompt in all 3 domains. Though our models suffered from some overfitting, particularly in the smallest sprite dataset, augmentation strategies were able to significantly improve generalization and metric performance. 

The 3 different datasets resulted in five-dollar models with varying properties. The map generator seemed to perform the best when given descriptive strings. This dataset consisted of sentence-long descriptions of maps, and we found the best results came from clearly describing scenes to generate.  

\begin{figure}[ht!]
    \centering
    \includegraphics[width=\linewidth]{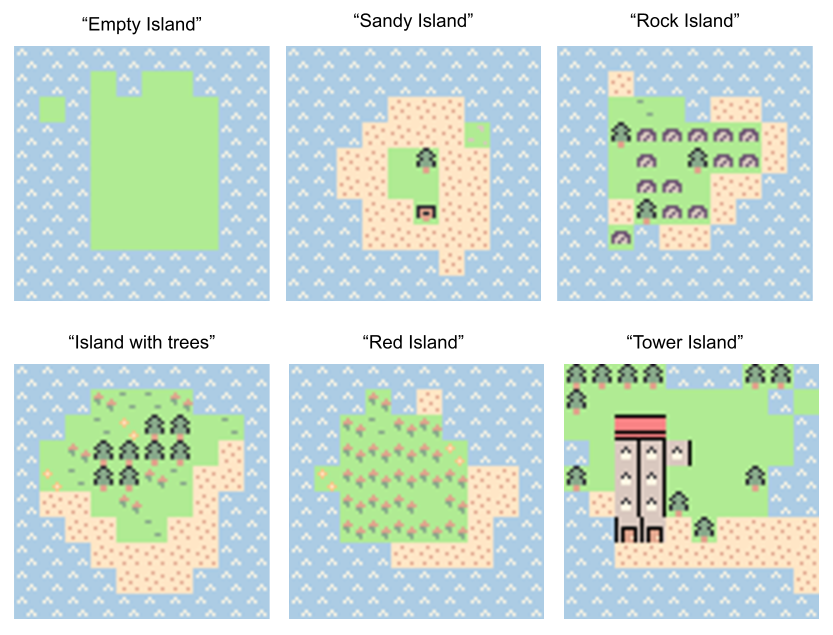}
    \caption{Results of the video game map generator for island generation. The model is able to understand concepts such as "empty", as well as tile names such as "tree" and "rock"}
    \label{fig:map_islands}
\end{figure}

\begin{figure}[ht!]
    \centering
    \includegraphics[width=\linewidth]{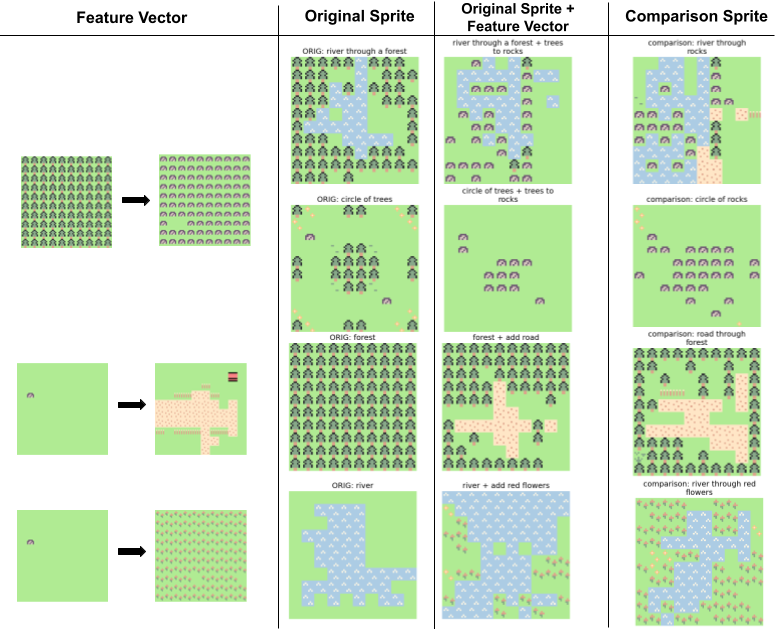}
    \caption{Feature Vector Arithmetic experiments in the Map domain. We use single word text prompts representing high-level map features, and experiment with applying them to other maps. The first two rows use a ``tree to rock" transformation. The third row use a ``road" vector to add a road. The fourth row uses a ``red flower" vector to add red flowers.
    }
    \label{fig:map_feature_vector}
\end{figure}

The map model also demonstrated a good understanding of the tileset. Text prompts that specify tiles, such as "trees", "rocks" and "water" were consistently accurate. We experiment with generating island maps with different features, depicted in Figure \ref{fig:map_islands}. Additionally, providing a single word text prompt such as "rocks" generated maps entirely covered by that tile. We utilize and explore this property in the vector arithmetic  experiments shown in Figure \ref{fig:map_feature_vector}.

\begin{figure}[ht!]
    \centering
    \includegraphics[width=\linewidth]{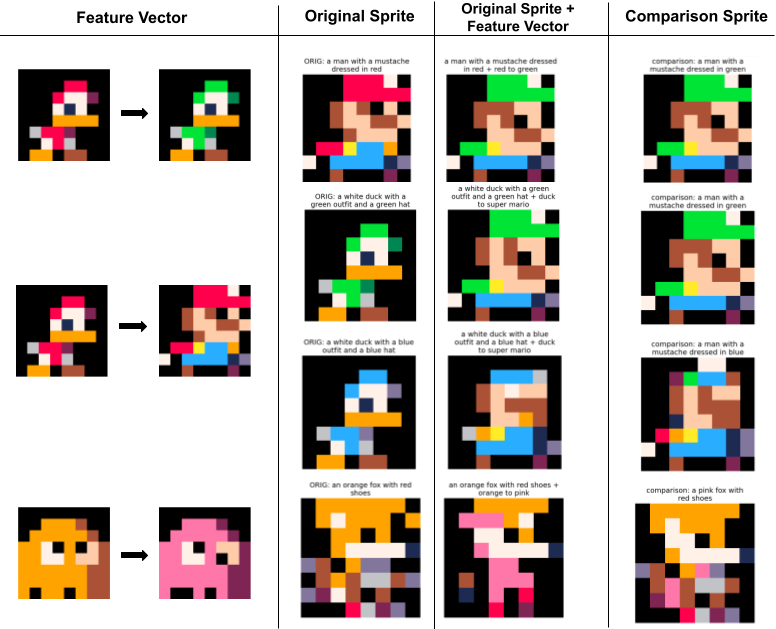}
    \caption{Vector Arithmetic experiments in the sprite domain. The first row isolates a ``red to green" feature vector, and applies it to Mario to get a Luigi sprite. The second and third rows use a ``duck to Mario Brother" vector. First we use it to transform a green duck into Luigi. Then, we apply it to a blue duck to obtain a novel blue Mario brothers character. The last row isolates a ``orange to pink" vector, then applies it to an orange fox. Notably, this method achieves a more accurate result for a pink fox than simply specifying ``pink" in the input text prompt.}
    \label{fig:sprite_feature_vector}
\end{figure}

The sprite generator was most effective in generating variations of existing characters, but understandably struggled to depict entirely original characters. 
This was the smallest of our datasets with only 100 samples, and these models often struggled with overfitting and memorization. The structure of this dataset (including similar characters with slight variations, such as color) gave this model the best understanding of color. We explore this in the vector arithmetic experiments shown in Figure \ref{fig:sprite_feature_vector}.

\begin{figure}[ht!]
    \centering
    \includegraphics[width=\linewidth]{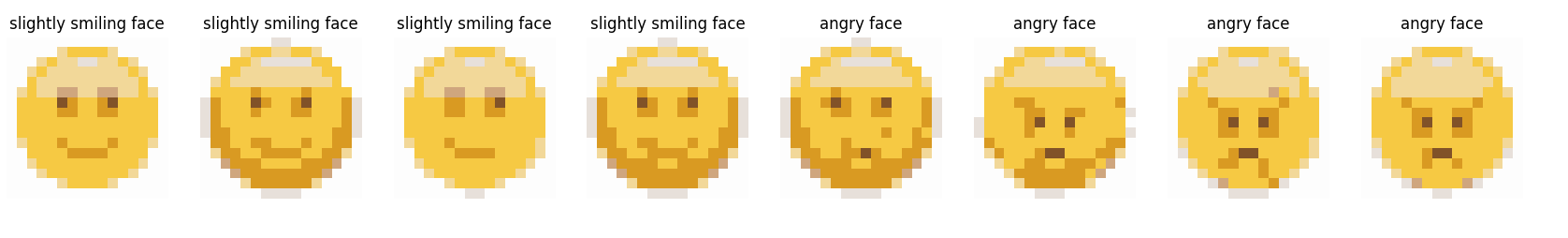}
    \includegraphics[width=\linewidth]{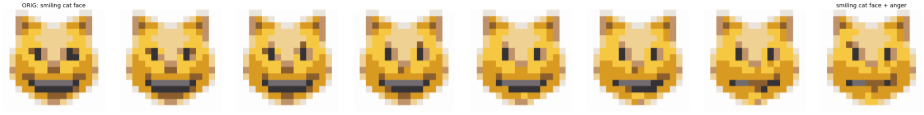}
    \caption{Interpolation and Feature Vector experiments in the Emoji domain. }
    \label{fig:feat_vec_interp}
\end{figure}

The emoji generator had the poorest performance in terms of generalization and generation of novel images. This dataset had the most diversity of images and subjects depicted. While emoji faces have a consistent structure and style, other categories like animals and flowers vary dramatically. While the model demonstrated good understanding of semantically similar descriptions (ex: Angry, Mad, or Furious face), it struggled to generate emoji faces depicting unseen emotions. Experiments on this model are shown in Figure \ref{fig:feat_vec_interp}. We tested models trained on only emoji faces, but did not notice much improvement. 

We find through interpolation experiments that the embedding space in all 3 domains is smooth, and results in smooth transitions between interpolated images. Additionally, we are able to perform vector arithmetic within the embedding space. This is demonstrated in Figures \ref{fig:map_feature_vector},
\ref{fig:sprite_feature_vector}, and 
\ref{fig:feat_vec_interp}.

\subsection{Grid Search for Best Architecture}
We find that within these 3 domains, deeper networks were not always better. The top performing models were consistently made up of 3 residual blocks, and achieved performance on par or exceeding the deeper 5 and 7 residual block architectures. In the sprite domain, deeper architectures (5 residual blocks) scored marginally higher in validation accuracy versus their 3 res block counterparts, though this difference was not noticeable in generated assets. We find the width of a network, i.e. a network with a larger kernel size or higher filter count, contributed the most to a models' generated image quality.

As seen in Table \ref{tab:appendix_table}, the experiments with more advanced methods of incorporating the textual condition, such as CIN and FiLM layers, received scores very similar to the simple concatenation-based input conditioning. From the CIN architecture we found slightly higher validation accuracy in the emojis domain with consistently better validation accuracy in the sprite domain. The map dataset, in comparison, had slightly worse accuracy with these large conditioning methods.

It was determined that these kinds of conditioning methods did not offer compelling enough improvements over simple concatenation-based input for the increase in the size of the architecture. As such, we leave the decision to the reader on which to implement based on the constraints of their downstream task.

\subsection{CLIP Metric}

The CLIP evaluation metric indicated positive results in terms of quality performance from our architectures. When analyzing both the emoji dataset and the sprite dataset, the CLIP model was able to easily differentiate between the real datasets and the fake, random datasets. This gave us a target CLIP score when evaluating our models, which generated well above the random level, and even up to the expected dataset level. The full data can be found in Table \ref{tab:data_clipscore}.

Despite preprompt testing, the CLIP model was unable to meaningfully score the real dataset above the random dataset for the map domain. We hypothesize that this is due to the map generation using pre-structured tiles (tiles that already represent information, such as a tree or a door), which could give enough structure to images for the CLIP model to s core the random images as high as the real data. The map generating models were still able to generate at a level consistent with the real data and noted subjectively high quality aesthetic output (as shown earlier in Figure \ref{fig:MainFigMaps}). This data is shown in Table \ref{tab:model_clipscore}.

\begin{figure}[ht!]
    \centering
    \includegraphics[width=\linewidth]{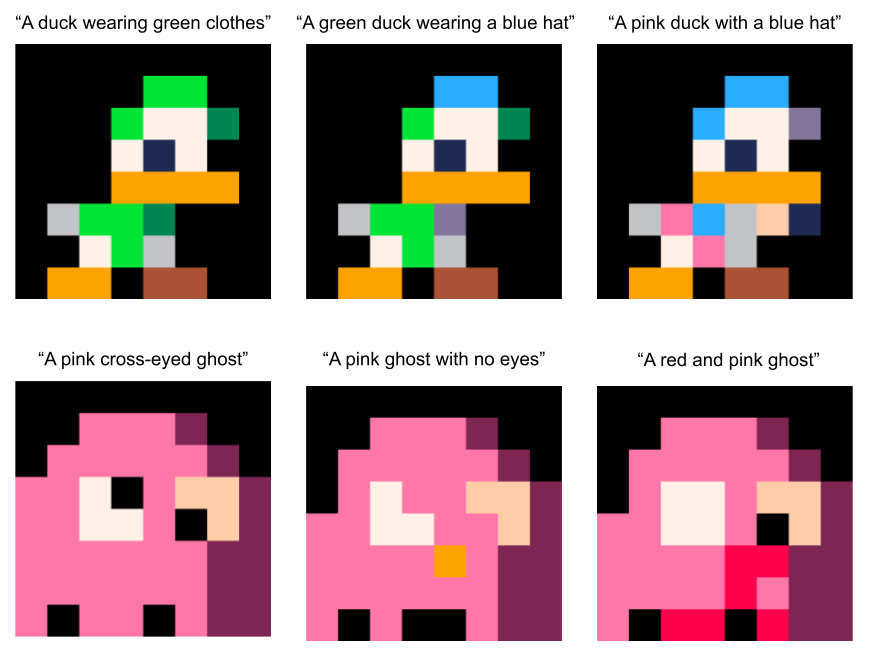}
    \caption{Results of the sprite generator on ducks and Ghosts. The closest labels in the dataset are "a white duck with a green outfit and a green hat", and "a pink cross-eyed ghost". The model is able to understand colors, and apply them to the shirt or hat of the duck sprites. It is also able to recognize and erase the pupils of the ghost sprite when given a caption with "no eyes"}
    \label{fig:sprite_ducks_ghosts}
\end{figure}

\section{Discussion}

We pursued this research topic as a means to have an alternative to generative adversarial networks and diffusion models for generating small content fast, especially when this content may not naturally be an RGB image. We especially focused on the lightweight aspect of the model as smaller models are more accessible. Having smaller models enables applications that larger models would not allow, such as real-time local generation. 
Exploring the model under the constraint of being much smaller than state-of-the-art alternatives also allows for easier reproducibility, iterative testing, and accessibility. 

Viewing these results through the lens of procedural game content generation it can be easy to imagine many possible use cases. Players could decide what the map, the characters, or any other asset looks like for infinite replayability. In the extreme, this could be done frame-to-frame with this new model's lightweight capabilities. This would allow for variability not just between user-prompted, generative playthroughs but to have individual characters with individual sprites, on tailor-made maps from any device all in real time.

That being said, it is important to realize the limitations of these new models. While this work demonstrates the capabilities of lightweight aesthetic generation in a novel, easily interpretable model, it may be the case that these architectures have been pushed close to their limit and could only have a limited amount of room to improve. The zero-shot capabilities of these models are also very limited as they are only able to generate pixel images related in some way to the training data, though possible future works could build on this issue to resolve it through clever prompting.

\section{Conclusion and Future Work}
This work has demonstrated the effectiveness of our simple, lightweight architecture for low-dimensional image generation with possible uses from video game maps to pixel art sprite generation. It also demonstrates the applicability of smaller models for specific use cases, even with limited data through the numerous augmentation methods discussed.   

Though we find that GPT-generated alternate labels and MixUp augmentations can improve the performance of our model in these small datasets, there are many more augmentation strategies we did not explore. NLP augmentation methods such as synonym replacement and back-translation may further improve the language understanding and generalization capabilities of our models. There may also be useful computer vision augmentation methods. Though we experimented with horizontal flips and image noising, these methods often introduced artifacts into generated images. Due to the categorical representation of our images (rather than the more common RGB), and extremely small image dimensions, many common image augmentation strategies are difficult or impossible to apply to our datasets.

We do not explore the use of attention mechanisms in our networks. Cross attention has shown incredible potential in text-to-image models, as demonstrated by models such as CAGAN \cite{10.1007/978-3-030-92659-5_25} and AttnGAN \cite{8578241}. Similar to CIN and FiLM, while attention may improve our performance, it also adds a significant amount of size and computational complexity to our models. We leave exploration of attention for categorical pixel image generation to future work.

\bibliography{ref.bib}

\begin{thebibliography}{24}
\providecommand{\natexlab}[1]{#1}

\bibitem[{Charity and Togelius(2022)}]{charity2022aesthetic}
Charity, M.; and Togelius, J. 2022.
\newblock Aesthetic Bot: Interactively Evolving Game Maps on Twitter.
\newblock In \emph{Proceedings of the AAAI Conference on Artificial
  Intelligence and Interactive Digital Entertainment}, volume~18, 18--25.

\bibitem[{Dumoulin, Shlens, and Kudlur(2017)}]{dumoulin2017a}
Dumoulin, V.; Shlens, J.; and Kudlur, M. 2017.
\newblock A Learned Representation For Artistic Style.
\newblock In \emph{International Conference on Learning Representations}.

\bibitem[{Earle et~al.(2022)Earle, Snider, Fontaine, Nikolaidis, and
  Togelius}]{earle2022illuminating}
Earle, S.; Snider, J.; Fontaine, M.~C.; Nikolaidis, S.; and Togelius, J. 2022.
\newblock Illuminating diverse neural cellular automata for level generation.
\newblock In \emph{Proceedings of the Genetic and Evolutionary Computation
  Conference}, 68--76.

\bibitem[{Hendrycks and Dietterich(2018)}]{hendrycks2018robustness}
Hendrycks, D.; and Dietterich, T. 2018.
\newblock Benchmarking Neural Network Robustness to Common Corruptions and
  Surface Variations.
\newblock \emph{arXiv preprint arXiv:1807.01697}.

\bibitem[{Hessel et~al.(2021)Hessel, Holtzman, Forbes, Le~Bras, and
  Choi}]{hessel-etal-2021-clipscore}
Hessel, J.; Holtzman, A.; Forbes, M.; Le~Bras, R.; and Choi, Y. 2021.
\newblock {CLIPS}core: A Reference-free Evaluation Metric for Image Captioning.
\newblock In \emph{Proceedings of the 2021 Conference on Empirical Methods in
  Natural Language Processing}, 7514--7528. Online and Punta Cana, Dominican
  Republic: Association for Computational Linguistics.

\bibitem[{Isola et~al.(2017)Isola, Zhu, Zhou, and Efros}]{isola2017image}
Isola, P.; Zhu, J.-Y.; Zhou, T.; and Efros, A.~A. 2017.
\newblock Image-to-image translation with conditional adversarial networks.
\newblock In \emph{Proceedings of the IEEE conference on computer vision and
  pattern recognition}, 1125--1134.

\bibitem[{Kang and Park(2020)}]{kang2020contragan}
Kang, M.; and Park, J. 2020.
\newblock Contragan: Contrastive learning for conditional image generation.
\newblock \emph{Advances in Neural Information Processing Systems}, 33:
  21357--21369.

\bibitem[{Mirza and Osindero(2014)}]{DBLP:journals/corr/MirzaO14}
Mirza, M.; and Osindero, S. 2014.
\newblock Conditional Generative Adversarial Nets.
\newblock \emph{CoRR}, abs/1411.1784.

\bibitem[{Perez et~al.(2017)Perez, Strub, de~Vries, Dumoulin, and
  Courville}]{Perez2017FiLMVR}
Perez, E.; Strub, F.; de~Vries, H.; Dumoulin, V.; and Courville, A.~C. 2017.
\newblock FiLM: Visual Reasoning with a General Conditioning Layer.
\newblock In \emph{AAAI Conference on Artificial Intelligence}.

\bibitem[{Radford et~al.(2021)Radford, Kim, Hallacy, Ramesh, Goh, Agarwal,
  Sastry, Askell, Mishkin, Clark, Krueger, and Sutskever}]{radford2021learning}
Radford, A.; Kim, J.~W.; Hallacy, C.; Ramesh, A.; Goh, G.; Agarwal, S.; Sastry,
  G.; Askell, A.; Mishkin, P.; Clark, J.; Krueger, G.; and Sutskever, I. 2021.
\newblock Learning Transferable Visual Models From Natural Language
  Supervision.
\newblock arXiv:2103.00020.

\bibitem[{Radford, Metz, and Chintala(2015)}]{radford2015unsupervised}
Radford, A.; Metz, L.; and Chintala, S. 2015.
\newblock Unsupervised representation learning with deep convolutional
  generative adversarial networks.
\newblock \emph{arXiv preprint arXiv:1511.06434}.

\bibitem[{Ramesh et~al.(2022)Ramesh, Dhariwal, Nichol, Chu, and
  Chen}]{ramesh2022hierarchical}
Ramesh, A.; Dhariwal, P.; Nichol, A.; Chu, C.; and Chen, M. 2022.
\newblock Hierarchical text-conditional image generation with clip latents.
\newblock \emph{arXiv preprint arXiv:2204.06125}.

\bibitem[{Reed et~al.(2016)Reed, Akata, Yan, Logeswaran, Schiele, and
  Lee}]{pmlr-v48-reed16}
Reed, S.; Akata, Z.; Yan, X.; Logeswaran, L.; Schiele, B.; and Lee, H. 2016.
\newblock Generative Adversarial Text to Image Synthesis.
\newblock In Balcan, M.~F.; and Weinberger, K.~Q., eds., \emph{Proceedings of
  The 33rd International Conference on Machine Learning}, volume~48 of
  \emph{Proceedings of Machine Learning Research}, 1060--1069. New York, New
  York, USA: PMLR.

\bibitem[{Saravanan and Guzdial(2022)}]{saravanan2022pixel}
Saravanan, A.; and Guzdial, M. 2022.
\newblock Pixel VQ-VAEs for Improved Pixel Art Representation.
\newblock \emph{arXiv preprint arXiv:2203.12130}.

\bibitem[{Sarkar and Cooper(2021)}]{sarkar2021generating}
Sarkar, A.; and Cooper, S. 2021.
\newblock Generating and blending game levels via quality-diversity in the
  latent space of a variational autoencoder.
\newblock In \emph{Proceedings of the 16th International Conference on the
  Foundations of Digital Games}, 1--11.

\bibitem[{Schulze, Yaman, and Waibel(2021)}]{10.1007/978-3-030-92659-5_25}
Schulze, H.; Yaman, D.; and Waibel, A. 2021.
\newblock CAGAN: Text-To-Image Generation with Combined Attention Generative
  Adversarial Networks.
\newblock In Bauckhage, C.; Gall, J.; and Schwing, A., eds., \emph{Pattern
  Recognition}, 392--404. Cham: Springer International Publishing.
\newblock ISBN 978-3-030-92659-5.

\bibitem[{Shaker, Togelius, and Nelson(2016)}]{pcgbook}
Shaker, N.; Togelius, J.; and Nelson, M.~J. 2016.
\newblock \emph{Procedural Content Generation in Games: A Textbook and an
  Overview of Current Research}.
\newblock Springer.

\bibitem[{Sinha et~al.(2021)Sinha, Song, Meng, and Ermon}]{sinha2021d2c}
Sinha, A.; Song, J.; Meng, C.; and Ermon, S. 2021.
\newblock D2c: Diffusion-decoding models for few-shot conditional generation.
\newblock \emph{Advances in Neural Information Processing Systems}, 34:
  12533--12548.

\bibitem[{Todd et~al.(2023)Todd, Earle, Nasir, Green, and
  Togelius}]{todd2023level}
Todd, G.; Earle, S.; Nasir, M.~U.; Green, M.~C.; and Togelius, J. 2023.
\newblock Level Generation Through Large Language Models.
\newblock In \emph{Proceedings of the 18th International Conference on the
  Foundations of Digital Games}, 1--8.

\bibitem[{Van~den Oord et~al.(2016)Van~den Oord, Kalchbrenner, Espeholt,
  Vinyals, Graves et~al.}]{van2016conditional}
Van~den Oord, A.; Kalchbrenner, N.; Espeholt, L.; Vinyals, O.; Graves, A.;
  et~al. 2016.
\newblock Conditional image generation with pixelcnn decoders.
\newblock \emph{Advances in neural information processing systems}, 29.

\bibitem[{Volz et~al.(2018)Volz, Schrum, Liu, Lucas, Smith, and
  Risi}]{volz2018evolving}
Volz, V.; Schrum, J.; Liu, J.; Lucas, S.~M.; Smith, A.; and Risi, S. 2018.
\newblock Evolving mario levels in the latent space of a deep convolutional
  generative adversarial network.
\newblock In \emph{Proceedings of the genetic and evolutionary computation
  conference}, 221--228.

\bibitem[{Xu et~al.(2023)Xu, Wang, Cheng, Cao, Shan, Qie, and
  Gao}]{xu2023dream3d}
Xu, J.; Wang, X.; Cheng, W.; Cao, Y.-P.; Shan, Y.; Qie, X.; and Gao, S. 2023.
\newblock Dream3d: Zero-shot text-to-3d synthesis using 3d shape prior and
  text-to-image diffusion models.
\newblock In \emph{Proceedings of the IEEE/CVF Conference on Computer Vision
  and Pattern Recognition}, 20908--20918.

\bibitem[{Xu et~al.(2018)Xu, Zhang, Huang, Zhang, Gan, Huang, and He}]{8578241}
Xu, T.; Zhang, P.; Huang, Q.; Zhang, H.; Gan, Z.; Huang, X.; and He, X. 2018.
\newblock AttnGAN: Fine-Grained Text to Image Generation with Attentional
  Generative Adversarial Networks.
\newblock In \emph{2018 IEEE/CVF Conference on Computer Vision and Pattern
  Recognition (CVPR)}, 1316--1324. Los Alamitos, CA, USA: IEEE Computer
  Society.

\bibitem[{Zhang et~al.(2017)Zhang, Cisse, Dauphin, and
  Lopez-Paz}]{zhang2017mixup}
Zhang, H.; Cisse, M.; Dauphin, Y.~N.; and Lopez-Paz, D. 2017.
\newblock mixup: Beyond empirical risk minimization.
\newblock \emph{arXiv preprint arXiv:1710.09412}.

\end{thebibliography}

\appendix

\newcolumntype{"}{@{\hskip\tabcolsep\vrule width 2pt\hskip\tabcolsep}}

\section{Appendix}

\begin{figure}[htbp]
    \centering
    \includegraphics[width=\linewidth]{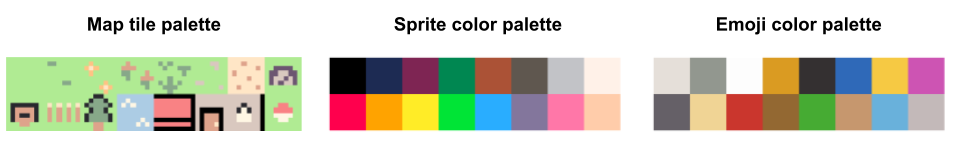}
    \caption{Dataset tile and color palettes}
    \label{fig:palettes}
\end{figure}

\lstset{
  basicstyle=\ttfamily,
  numbers=none,          
  frame=single,          
  breaklines=true,         
  breakindent=10pt,
  aboveskip=1em          
}

\begin{table}[htbp]
\centering
    \begin{tabular}{|c"c|c|}
    \hline
    Domain & Dataset Type & Average CLIP score \\
    \hline \hline
    Maps & Real Data & $0.760$ \\
    \hline
    Maps & Random Data & $0.750$ \\
    \hline \hline
    Sprites & Real Data & $0.803$ \\
    \hline
    Sprites & Random Data & $0.638$ \\
    \hline \hline
    Emojis & Real Data & $0.800$ \\
    \hline
    Emojis & Random Data & $0.705$ \\
    \hline
    \end{tabular}
    \caption{The Average CLIP score for real labels with Real image data and the real labels with Random image data in each domain}
    \label{tab:data_clipscore}
\end{table}

\begin{table}[htbp]
    \centering
    \begin{tabular}{|c"c|c|}
        \hline
        Domain & Conditioning Type & CLIP Score on Unseen \\
        \hline 
        \hline
        Maps & Standard & $0.758$ \\
        \hline
        Maps & FiLM & $0.765$ \\ 
        \hline
        \hline
        Sprites & Standard & $0.813$ \\
        \hline
        Sprites & CIN & $0.830$ \\
        \hline 
        \hline
        Emojis & Standard & $0.715$ \\
        \hline
        Emojis & CIN & $0.743$ \\
        \hline
    \end{tabular}
    \centering
    \caption{The average CLIP score of each of the top performing standard and non-standard models in each domain}
    \label{tab:model_clipscore}
\end{table}

\begin{table*}[htbp]
  \centering
  \begin{tabular}{|c"c|c|c|c|c"c|}
\hline
Domain & Noise Dimension & Filter Count & Kernel Size & Res Blocks & Conditioning Type & Highest Validation Accuracy\\
\hline\hline
Maps & 5 & 256 & 7 & 3 & Standard & $0.854$\\
\hline
Maps & 5 & 256 & 7 & 3 & FiLM & $0.849$\\
\hline\hline
Sprites & 5 & 256 & 7 & 5 & CIN & $0.997$\\
\hline
Sprites & 5 & 256 & 7 & 6 & Standard & $0.943$\\
\hline\hline
Emojis & 5 & 256 & 7 & 3 & CIN & $0.829$\\
\hline
Emojis & 5 & 256 & 5 & 3 & Standard & $0.812$\\
\hline
\end{tabular}
  \caption{Grid Search results with best performing standard model and experimental conditioning models separated by each domain.}
  \label{tab:appendix_table}
\end{table*}

\begin{figure*}
\begin{lstlisting}
Take each string in the list and write an alternate label for each one.These strings describe an image of a pixel video game map. These alternate labels should describe the same image as the original label, but use different words and a different sentence structure. Use simple or common words when writing the alternate labels. Assume you have the vocabulary of a 10 year old. Your output should have the same number of strings as the input list.
\end{lstlisting}
\caption{The GPT prompt for the Map dataset. Prompts for the Sprite and Emoji datasets were similar, but replaced "pixel video game map"}
\label{lst:gpt_prompt}
\end{figure*}

\begin{figure*}
\begin{lstlisting}
newunseen_map = ["a house with a pool in the forest", "a flower garden by the beach", "a trail through a seaside town", "a lake in the middle of a forest town", "rows of flowers by a house in a desert"]

newunseen_emoji = ["hand with three fingers up", "a face with a wide smile", "a face raising one eyebrow", "sad cat face", "a devil face with sunglasses"]

newunseen_sprite = ["a buff man with a blue headband and red shirt", "a blue duck with a red headband", "a green woman with blonde hair", "a dog with a black hat", "a man with blue shoes, a red hat, and a green shirt"]

\end{lstlisting}
\caption{Examples of unseen prompts for each domain}
\label{lst:unseen_listings}
\end{figure*}

\begin{figure*}
\begin{lstlisting}
preprompt_map = "a frame from a pixel game map of "
preprompt_emoji = "a pixelated emoji of "
preprompt_sprite = "a pixel art style sprite of "
\end{lstlisting}
\caption{The preprompts used for each domain type}
\label{lst:preprompt_listings}
\end{figure*}

\end{document}